\documentclass[11pt,a4paper]{article}
\usepackage[hyperref]{emnlp2020}
\usepackage{times}
\usepackage{latexsym}

\usepackage{soul}
\usepackage{url}
\usepackage{algorithm}
\usepackage{algorithmicx}
\usepackage[noend]{algpseudocode}
\usepackage{balance}
\newcommand{\ModelName}{\textsc{CLANG}}
\usepackage{amsmath,amsthm,mathtools}
\usepackage{latexsym,graphicx,amssymb,multirow,booktabs}
\usepackage{makecell}

\usepackage{subfigure}
\usepackage[font={footnotesize}]{caption}

\usepackage{hyperref}
\usepackage{url}
\usepackage{xcolor}
\usepackage{soul}
\usepackage{enumitem}
\usepackage{mathtools}

\usepackage{microtype}

\aclfinalcopy % Uncomment this line for the final submission
\def\aclpaperid{2487} %  Enter the acl Paper ID here

\newcommand{\KL}{\mathbb{D}_{\rm{KL}}}

\title{Composed Variational Natural Language Generation for Few-shot Intents}
\author{Congying Xia$^{1}$\thanks{~ Work was done when Congying was a research intern at Salesforce Research.}, Caiming Xiong$^{2}$, Philip Yu$^{1}$ and Richard Socher$^{2}$\\
{$^1$University of Illinois at Chicago, Chicago, IL, USA} \\
{$^2$Salesforce Research, Palo Alto, CA, US} \\
 {\tt \{cxia8, psyu\}@uic.edu},
 {\tt \{cxiong, rsocher\}@salesforce.com}
 }
\date{}

\begin{document}
\maketitle
\begin{abstract}
In this paper, we focus on generating training examples for few-shot intents in the realistic imbalanced scenario. To build connections between existing many-shot intents and few-shot intents, we consider an intent as a combination of a domain and an action, and propose a composed variational natural language generator (CLANG), a transformer-based conditional variational autoencoder. CLANG utilizes two latent variables to represent the utterances corresponding to two different independent parts (domain and action) in the intent, and the latent variables are composed together to generate natural examples. Additionally, to improve the generator learning, we adopt the contrastive regularization loss that contrasts the in-class with the out-of-class utterance generation given the intent. To evaluate the quality of the generated utterances, experiments are conducted on the generalized few-shot intent detection task. Empirical results show that our proposed model achieves state-of-the-art performances on two real-world intent detection datasets.

\end{abstract}
\section{Introduction}
Intelligent assistants have gained great popularity in recent years since they provide a new way for people to interact with the Internet conversationally \citep{hoy2018alexa}. However, it is still challenging to answer people's diverse questions effectively. Among all the challenges, identifying user intentions from their spoken language is important and essential for all the downstream tasks.

Most existing works \citep{hu2009understanding, xu2013convolutional, chen2016end, xia2018zero} formulate intent detection as a classification task and achieve high performance on pre-defined intents with sufficient labeled examples. With this ever-changing world, a realistic scenario is that we have imbalanced training data with existing many-shot intents and insufficient few-shot intents. Previous intent detection models  \cite{yin2020meta, yin2019benchmarking} deteriorate drastically in discriminating the few-shot intents. 

To alleviate this scarce annotation problem, several methods \citep{wei2019eda, malandrakis2019controlled, yoo2019data} have proposed to augment the training data for low-resource spoken language understanding (SLU). \citet{wei2019eda} introduce simple data augmentation rules for language transformation like insert, delete and swap. \citet{malandrakis2019controlled} and \citet{yoo2019data} utilize variational autoencoders \cite{kingma2013auto} with simple LSTMs \citep{hochreiter1997long} that have limited model capacity to do text generation. Furthermore, these models are not specifically designed to transfer knowledge from existing many-shot intents to few-shot intents.

In this paper, we focus on transferable natural language generation by learning how to compose utterances with many-shot intents and transferring to few-shot intents. When users interact with intelligent assistants, their goal is to query some information or execute a command in a certain domain \cite{ibm2017doc}. For instance, the intent of the input ``what will be the highest temperature next week'' is to ask about the weather. The utterance can be decomposed into two parts, ``what will be'' corresponding to an action ``Query'' and ``the highest temperature'' related to the domain ``Weather''. These actions or domains are very likely to be shared among different intents including the few-shot ones \cite{xu2019open}. For example, there are a lot of actions (``query'', ``set'', ``remove'') that can be combined with the domain of ``alarm''. The action ``query'' also exists in multiple domains like ``weather'', ``calendar'' and ``movie''. Ideally, if we can learn the expressions representing for a certain action or domain and how they compose an utterance for existing intents, then we can learn how to compose utterances for few-shot intents naturally. Therefore, we define an intent as a combination of a domain and an action. Formally, we denote the domain as $y_d$ and the action as $y_a$. Each intent can be expressed as $y = (y_d, y_a)$.

A composed variational natural language generator (CLANG) is proposed to learn how to compose an utterance for a given intent with an action and a domain. CLANG is a transformer-based \citep{vaswani2017attention} conditional variational autoencoder (CVAE) \citep{NIPS2014_5352}. It contains a bi-latent variational encoder and a decoder. The bi-latent variational encoder utilizes two independent latent variables to model the distributions of action and domain separately.
%The bi-latent component also improves the model's flexibility.
Special attention masks are designed to guide these two latent variables to focus on different parts of the utterance and disentangle the semantics for action and domain separately. Through decomposing utterances for existing many-shot intents, the model learns to generate utterances for few-shot intents as a composition of the learned expressions for domain and action.

Additionally, we adopt the contrastive regularization loss to improve our generator learning. 
During the training, an in-class utterance from one intent is contrasted with an out-of-class utterance from another intent. Specifically, the contrastive loss is to constrain the model to generate the positive example with a higher probability than the negative example with a certain margin. With the contrastive loss, the model is regularized to focus on the given domain and intent and the probability of generating negative examples is reduced.

To quantitatively evaluate the effectiveness of CLANG for augmenting training data in low-resource intent detection, experiments are conducted for the generalized few-shot intent detection task (GFSID) \cite{xia2020cg}. GFSID aims to discriminate a joint label space consisting of both existing many-shot intents and few-shot intents. 

Our contributions are summarized below. 1) We define an intent as a combination of a domain and an action to build connections between existing many-shot intents and few-shot intents. 2) A composed variational natural language generator (CLANG) is proposed to learn how to compose an utterance for a given intent with an action and a domain. Utterances are generated for few-shot intents via a composed variational inferences process. 3) Experiment results show that {\ModelName} achieves state-of-the art performance on two real-world intent detection datasets for the GFSID task.

\begin{figure*}[ht!]
    \centering
    \includegraphics[width=\linewidth]{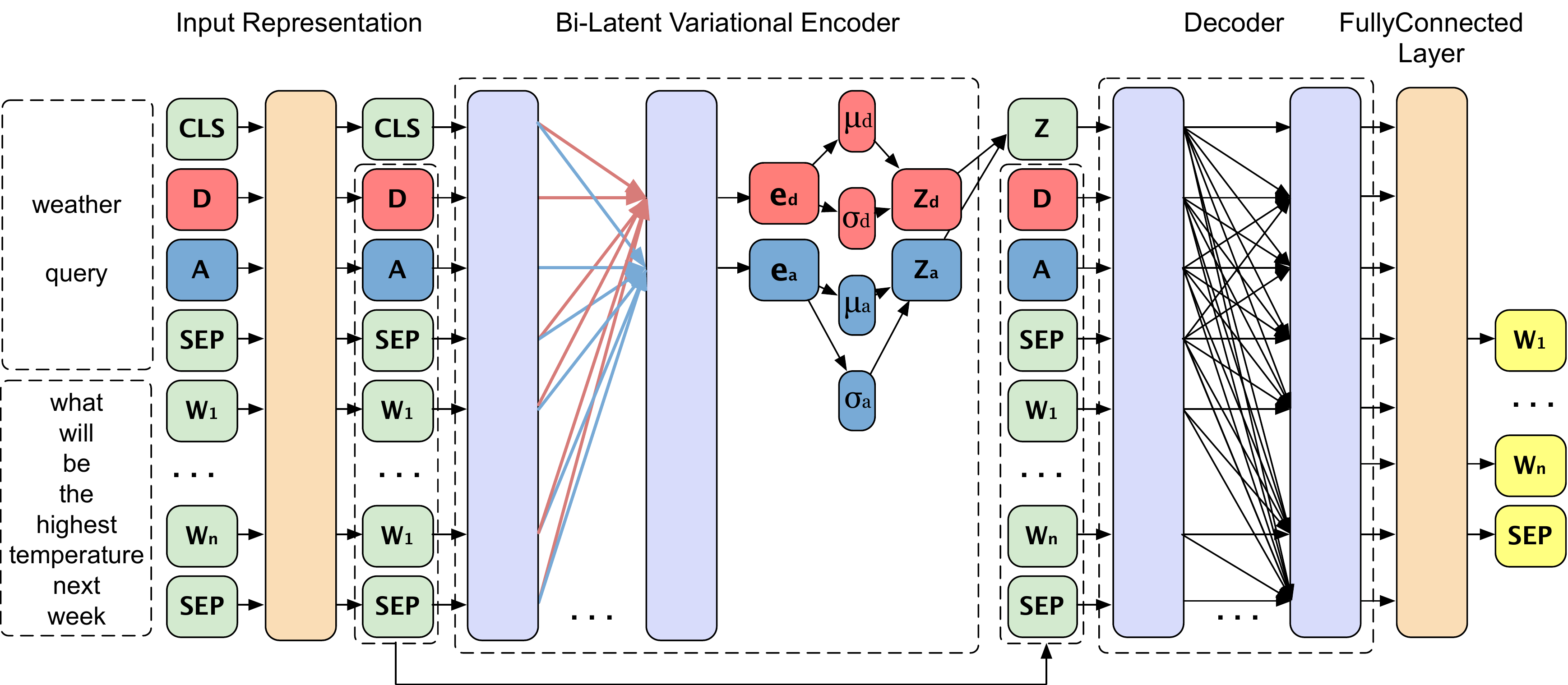}
    %\vspace{-0.2in}
    \caption{The overall framework of {\ModelName}.}
    \vspace{-0.1in}
    \label{fig:framework}
\end{figure*}

%\vspace{-0.05in}
\section{Composed Variational Natural Language Generator}
%\vspace{-0.05in}
In this section, we introduce the composed variational natural language generator (CLANG). As illustrated in Figure \ref{fig:framework}, CLANG consists of three parts: input representation, bi-latent variational encoder and decoder.

%\vspace{-0.05in}
\subsection{Input Representation}
%\vspace{-0.05in}
\label{input}
For a given intent $y$ decomposed into a domain $y_d$ and an action $y_a$ and an utterance $x = (w_1, w_2, ..., w_n)$ with $n$ tokens, we designed the input format like BERT as  ([CLS], $y_d$, $y_a$, [SEP], $w_1$, $w_2$, ..., $w_n$, [SEP]). As the example in Figure 1, the intent has the domain of ``weather'' and the action of ``query''. The utterance is ``what will be the highest temperature next week''. The input is represented as ([CLS], weather, query, [SEP], what, will, be, the, highest, temperature, next, week, [SEP]).

Texts are tokenized into subword units by WordPiece \cite{wu2016google}. The input embeddings of a token sequence are represented as the sum of three embeddings: token embeddings, position embeddings \cite{vaswani2017attention}, and segment embeddings \cite{devlin2018bert}. The segment embeddings are learned to identify the intent and the utterance with different embeddings.

\subsection{Bi-latent Variational Encoder}
\label{Encoder}
As illustrated in Figure \ref{fig:framework}, the bi-latent variational encoder is to encode the input into two latent variables that contain the disentangled semantics in the utterance corresponding to domain and action separately. 

Multiple transformer layers \cite{vaswani2017attention} are utilized in the encoder. 
Through the self-attention mechanism, these transformer layers not only extract semantic meaningful representations for the tokens, but also model the relation between the intent and the utterance. The embeddings for the domain token and the action token output from the last transformer layer are denoted as $\mathbf{e}_d$ and $\mathbf{e}_a$. We encode $\mathbf{e}_d$ into variable $z_d$ to model the distribution for the domain and $\mathbf{e}_a$ are encoded into variable $z_a$ to model the distribution for the action. 

Ideally, we want to disentangle the information for the domain and the action, making $\mathbf{e}_d$ attend to tokens related to domain and $\mathbf{e}_a$ focus on the expressions representing the action. To achieve that, we make a variation of the attention calculations in transformer layers to avoid direct interactions among the domain token and the action token in each layer.

Instead of applying the whole bidirectional attention to the input, an attention mask matrix $\mathbf{M} \in \mathbb{R}^{N \times N}$ is added to determine whether a pair of tokens can be attended to each other \cite{dong2019unified}. $N$ is the length of the input. For the $l$-th Transformer layer, the output of a self-attention head $\mathbf{A}_l$ is computed via:
%\vspace{-0.1in}
\begin{align}
\begin{aligned}
    \mathbf{Q} &= \mathbf{T}^{l-1}\mathbf{W}^l_Q, \\
    \mathbf{K} &= \mathbf{T}^{l-1}\mathbf{W}^l_K, \\
    \mathbf{V} &= \mathbf{T}^{l-1}\mathbf{W}^l_V, \\
    {\mathbf{A}_l} &= \text{softmax} \left(\frac{\mathbf{Q}\mathbf{K}^{\top}}{\sqrt {d_k}}  + \mathbf{M} \right){\mathbf{V}}, 
    \end{aligned}
%    \vspace{-0.1in}
\end{align}
%\vspace{-0.02in}
where the attention mask matrix calculated as: \begin{equation}
%\vspace{-0.05in}
\mathbf{M}_{ij} =
    \begin{cases}
      0, & \text{allow to attend;}\\
      -\infty,  & \text{prevent from attending.}
    \end{cases}
\end{equation}
The output of the previous transformer layer $\mathbf{T}^{l-1}{\in}\mathbb{R}^{N \times d_h}$ is linearly projected to a triple of queries, keys and values parameterized by matrices $\mathbf{W}^l_Q, \mathbf{W}^l_K, \mathbf{W}^l_V{\in}\mathbb{R}^{d_h{\times}d_k}$. $d_h$ is the hidden dimension for the transformer layer and $d_k$ is the hidden dimension for a self-attention head.

\begin{figure}[ht!]
    \centering
    \includegraphics[width=\linewidth]{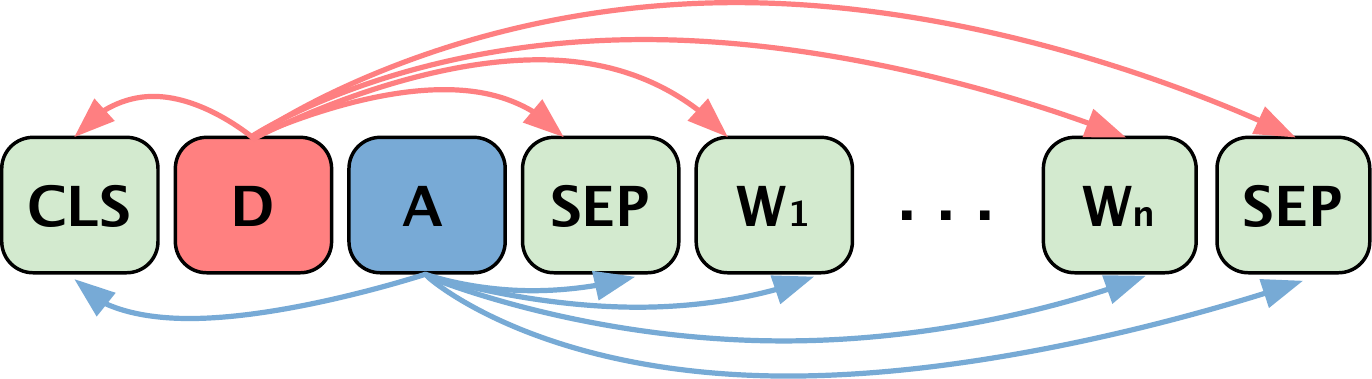}
   \vspace{-0.1in}
    \caption{The attention map of domain and action in the encoder.}
    \vspace{-0.1in}
    \label{fig:mask}
\end{figure}

The proposed attention mask for the domain token and the action token is illustrated in Figure \ref{fig:mask}. The Domain $y_d$ and the action $y_a$ are prevent from attending to each other. All the other tokens have are allowed to have full attentions. The elements in the mask matrix for the attentions between domain and action are $-\infty$, and 0 for all the others.

The disentangled embeddings $\mathbf{e}_d$ and $\mathbf{e}_a$ are encoded into two latent variables $z_d$ and $z_a$ to model the posterior distributions determined by the intent elements separately: $p(z_d|x, y_d)$, $p(z_a|x, y_a)$. The latent variable $z_d$ is conditioned on the domain $y_d$, while $z_a$ is controlled by the action $y_a$. By modeling the true distributions, $p(z_d|x, y_d)$ and $p(z_a|x, y_a)$, using a known distribution that is easy to sample from \citep{NIPS2014_5352}, we constrain the prior distributions, $p(z_d|y_d)$ and $p(z_a|y_a)$, as multivariate standard Gaussian distributions. A reparametrization trick \cite{kingma2013auto} is used to generate the latent vector $z_d$ and $z_a$ separately.
Gaussian parameters ($\mu_d$, $\mu_a$, $\sigma_d ^2$, $\sigma_a ^2$) are projected from $\mathbf{e}_d$ and $\mathbf{e}_a$:
%\vspace{-0.1in}
\begin{align}
	\begin{aligned}
 %   \vspace{-0.1in}
	\mu_d  &= \mathbf{e}_d{\textbf{W}_{\mu_d} } + {b_{\mu_d} }, \\
    {\log(\sigma_d ^2)} &= \mathbf{e}_d{\textbf{W}_{\sigma_d} } + {b_{\sigma_d} }, \\
    \mu_a  &= \mathbf{e}_a{\textbf{W}_{\mu_a} } + {b_{\mu_a} }, \\
    {\log(\sigma_a ^2)} &= \mathbf{e}_a{\textbf{W}_{\sigma_a} } + {b_{\sigma_a} }, 
    %\vspace{-0.1in}
	\end{aligned}
\end{align}
%\vspace{-0.02in}
where we have $\textbf{W}_{\mu_d}, \textbf{W}_{\mu_a}, \textbf{W}_{\sigma_d}, \textbf{W}_{\sigma_a} \in \mathbb{R}^{d_h \times d_h}$, $b_{\mu_d}, b_{\mu_a}, b_{\sigma_d}, b_{\sigma_a} \in \mathbb{R}^{d_h}$. Noisy variables $\varepsilon_d \sim \mathcal{N}(0, \mathrm{I}), \varepsilon_a \sim \mathcal{N}(0, \mathrm{I})$ are utilized to sample $z_d$ and $z_a$ from the learned distribution:
%\vspace{-0.05in}
\begin{align}
	\begin{aligned}
%    \vspace{-0.1in}
	{z_d} &= \mu_d + {\sigma_d}\cdot \varepsilon_d, \\
	{z_a} &= \mu_a + {\sigma_a} \cdot \varepsilon_a.
%    \vspace{-0.1in}
	\end{aligned}
\end{align}

The KL-loss function is applied to regularize the prior distributions for these two latent variables to be close to the Gaussian distributions:
\begin{align}
	\begin{aligned}
    \mathcal{L}_{KL}& = \KL [ {q( {z_d|x,y_d}), p(z_d|y_d)}] \\
   &+ \KL [ {q(z_a|x, y_a), p(z_a|y_a)}]\,.
%    \vspace{-0.1in}
	\end{aligned}
\end{align}

A fully connected layer with Gelu \cite{hendrycks2016bridging} activation function is applied on $z_d$ and $z_a$ to compose these two latent variables together and outputs $z$. The composed latent information $z$ is utilized in the decoder to do generation.

%\vspace{-0.05in}
\subsection{Decoder}
%\vspace{-0.05in}
\label{Decoder}
The decoder utilizes the composed latent information together with the intent to reconstruct the input utterance $p(x|z_d, z_a, y_d, y_a)$. As shown in Figure \ref{fig:framework}, a residual connection is built from the input representation to the decoder to get the embeddings for all the tokens. To keep a fixed length and introduce the composed latent information $z$ into the decoder, we replace the first [CLS] token with $z$.

The decoder is built with multiple transformer layers to generate the utterance. Text generation is a sequential process that we use the left context to predict the next token. To simulate the left-to-right generation process, another attention mask is utilized for the decoder. In the attention mask for the decoder, tokens in the intent can only attend to intent tokens, while tokens in the utterance can attend to both the intent and all the left tokens in the utterance.

For the first token $z$ which holds composed latent information, it is only allowed to attend to itself due to the vanishing latent variable problem.
The latent information can be overwhelmed by the information of other tokens when adapting VAE to natural language generators either for LSTM \citep{zhao2017learning} or transformers \cite{xia2020cg}. To further increase the impact of the composed latent information $z$ and alleviate the vanishing latent variable problem, we concatenate the token representations of $z$ to all the other token embeddings output from the last transformer layer in the decoder.

The hidden dimension increases to $2 \times d_h$ after the concatenation. To reduce the hidden dimension to $d_h$ and get the embeddings to decode the vocabulary, two fully-connected (FC) layers followed by a layer normalization \cite{ba2016layer} are applied on top of the transformer layers. Gelu is used as the activation function in these two FC layers. The embeddings output from these two FC layers are decoded into tokens in the vocabulary. The embeddings at position $i = \{1, ..., n-1\}$ are used to predict the next token at position $i+1$ till the [SEP] token is generated.

To train the decoder to reconstruct the input, a reconstruction loss is formulated as:
\begin{align}
   \mathcal{L}_r=-{\mathbb{E}_{q(z_d|x,y_d),q(z_a|x, y_a)}}[ {\log p( {x|z_d, z_a,y_d, y_a})}]. 
    \vspace{-0.1in}
\end{align}

\vspace{-0.05in}
\subsection{Learning with contrastive loss}
%\vspace{-0.05in}
\label{contrastive}
Although the model can generate utterances for a given intent, such as ``are there any alarms set for seven am'' for ``Alarm Query'', there are some negative utterances generated. For example, ``am i free between six to seven pm'' is generated with the intent of ``Alarm Query''. This would be because in the training, it lacks supervision to distinguish in-class from out-of-class examples especially for few-shot intents. To alleviate the problem, we adopt a contrastive loss in the objective function and reduce the probability to generate out-of-class samples.

Given an intent $y = (y_d, y_a)$, an in-class utterance $x^+$ from this intent and an out-of-class utterance $x^-$ from another intent. The contrastive loss constrains the model to generate the in-class example $x^+$ with a higher probability than $x^-$. In the same batch, we feed the in-class example ($y_d, y_a, x^+$) and the out-of-class example ($y_d, y_a, x^-$) into {\ModelName} to model the likelihood: $P(x^+|y)$ and $P(x^-|y)$. The chain rule is used to calculate the likelihood of the whole utterance: $p(x|y) = p(w_1|y)p(w_2|y, w_1)...p(w_n|y, w_1, ..., T_{w_{n-1}})$. In the contrastive loss, the log-likelihood of the in-class example is constrained to be higher than the out-of-class example with a certain margin $\lambda$:
% \vspace{-0.05in}
\begin{equation}
%    \vspace{-0.05in}
     \mathcal{L}_{c} = \text{max}\{0, \lambda - \text{log}p(x^+|y) + \text{log}p(x^- |y)\}.
%    \vspace{-0.02in}
     \label{eq:contrastive_loss}
\end{equation}

To leverage challenging out-of-class utterances, we choose the most similar utterance with a different intent as the out-of-class utterance. Three indicators are considered to measure the similarity between the in-class utterance and all the utterances with a different intent: the number of shared uni-grams $s_1$, bi-grams $s_2$ between the utterances and the number of shared uni-grams between the name of intents $s_3$. The sum of these three numbers, $s = s_1 + s_2 + s_3$, is utilized to find the out-of-class utterance with the highest similarity. If there are multiple utterances having the same highest similarity $s$, we random choose one as the negative utterance.

The overall loss function is a summation of the KL-loss, the reconstruction loss and the contrastive loss:
\vspace{-0.2in}
\begin{align}
\vspace{-0.1in}
   \mathcal{L}& =  \mathcal{L}_{KL} + \mathcal{L}_r + \mathcal{L}_c\,.
   \vspace{-0.1in}
\end{align}

\subsection{Generalized Few-shot Intent Detection}
Utterances for few-shot intents are generated by sampling two latent variables, $z_d$ and $z_a$, separately from multivariate standard Gaussian distributions. Beam search is applied to do the generation. To improve the diversity of the generated utterances, we sample the latent variables for $s$ times and save the top $k$ results for each time. The overall generation process follows that of \citet{xia2020cg}.

These generated utterances are added to the original traning dataset to alleviate the scare annotation problem. We finetune BERT with the augmented dataset to solve the generalized few-shot intent detection task. The whole pipeline is referred as BERT + {\ModelName}  in the experiments.
%\vspace{-0.05in}
\begin{table}[ht!]
\centering
\resizebox{\linewidth}{!}{% <-
\begin{tabular}{l|c|c}
\Xhline{3\arrayrulewidth}
 Dataset & SNIPS-NLU & NLUED \\ \hline
Vocab Size & 10,896 & 6,761\\ 
\#Total Classes & 7 & 64 \\ 
\#Few-shot Classes & 2 & 16\\
\#Few-shots / Class & 1 or 5 & 1 or 5 \\
\#Training Examples & 7,858 & 7,430\\ 
\#Training Examples / Class & 1571.6 & 155\\
\#Test Examples & 2,799 & 1,076\\ 
Average Sentence Length   &  9.05    & 7.68\\
\Xhline{3\arrayrulewidth}
\end{tabular}
}
%\vspace{-0.05in}
\caption{Data Statistics for SNIPS-NLU and NLUED. \#Few-shot examples are excluded in the \#Training Exampels. For NLUED, the statistics is reported for KFold\_1.}
\vspace{-0.1in}
\label{dataset}
\end{table}

\begin{table*}[ht!]
\centering
\resizebox{\linewidth}{!}{% <-
\begin{tabular}{l|ccc|ccc}
\Xhline{5\arrayrulewidth}
& Many-shot  & Few-shot & H-Mean  & Many-shot  & Few-shot & H-Mean \\ 
                     & \multicolumn{3}{c|}{SNIPS-NLU 1-shot}     & \multicolumn{3}{c}{SNIPS-NLU 5-shot}    \\ \Xhline{3\arrayrulewidth}
BERT-PN+ &92.66 $\pm$ 4.49 & 60.52 $\pm$ 7.58 &	72.99 $\pm$ 5.97  &	95.96 $\pm$ 1.13&	86.03 $\pm$	2.00&	90.71 $\pm$ 1.19  \\ 
BERT & 98.20 $\pm$ 0.06& 44.42 $\pm$ 4.35 & 57.74 $\pm$ 7.50  & 98.34 $\pm$ 0.10& 81.82 $\pm$ 6.16 &89.22 $\pm$ 3.74  \\
BERT + SVAE &98.24 $\pm$ 0.09 & 45.15 $\pm$ 5.54& 61.67 $\pm$ 5.11& \textbf{98.34} $\pm$ \textbf{0.06} & 82.10 $\pm$ 4.06&	89.49 $\pm$ 2.47  \\
BERT + CGT  & 98.20 $\pm$ 0.07&45.80 $\pm$ 5.68&62.30 $\pm$ 5.17&98.32 $\pm$ 0.14&82.65 $\pm$ 4.31&89.78 $\pm$ 2.83\\
BERT + EDA & 98.20 $\pm$ 0.08& 47.52 $\pm$ 5.96& 63.87 $\pm$ 5.29& 98.09 $\pm$ 0.18 &82.00 $\pm$ 3.47&89.30 $\pm$ 2.12\\
BERT + CG-BERT   
& 98.13 $\pm$ 0.15 &63.04 $\pm$ 5.49 & 76.65 $\pm$ 4.24 
& 98.30 $\pm$ 0.17& 86.89 $\pm$ 4.05 & 92.20 $\pm$ 2.32\\
\Xhline{3\arrayrulewidth}
{BERT + {\ModelName}} &\textbf{98.34 $\pm$ 0.10}&\textbf{64.63 $\pm$ 6.16}& \textbf{77.86 $\pm$ 4.39}& \textbf{98.34 $\pm$ 0.06} & \textbf{88.04 $\pm$ 1.34} &\textbf{92.90 $\pm$ 0.71}  \\
\Xhline{5\arrayrulewidth}
                     & \multicolumn{3}{c|}{NLUED 1-shot}     & \multicolumn{3}{c}{NLUED 5-shot}  
                      \\ \Xhline{3\arrayrulewidth}
BERT-PN+ &	81.24 $\pm$ 2.76&	18.95 $\pm$ 4.42&	30.67 $\pm$ 5.53 &83.41 $\pm$ 2.62&	60.28 $\pm$ 4.19&	69.93 $\pm$ 3.49\\  
BERT        
& 94.00 $\pm$ 0.93 & 7.88 $\pm$ 3.28 & 14.39 $\pm$ 5.66  &\textbf{94.12 $\pm$ 0.89}& 51.69 $\pm$ 3.19 &	66.67 $\pm$ 2.51  \\
BERT + SVAE & 93.80 $\pm$ 	0.70&  8.88 $\pm$ 3.66& 16.01 $\pm$ 6.06 & 93.60 $\pm$ 	0.63&	54.03 $\pm$ 3.91 &  68.42 $\pm$ 3.06 \\
BERT + CGT &94.00 $\pm$ 0.66& 9.33 $\pm$ 3.68&16.78 $\pm$ 6.16&93.61 $\pm$ 0.63&54.70 $\pm$ 4.06&68.96 $\pm$ 3.17   \\
BERT + EDA  &93.78 $\pm$ 0.66 & 11.65 $\pm$ 4.89& 20.41 $\pm$ 7.56&93.71 $\pm$ 0.64&57.22 $\pm$ 4.35 &70.95 $\pm$ 3.35\\
BERT + CG-BERT &\textbf{94.01 $\pm$ 0.70} & 20.39 $\pm$ 5.77 &33.12 $\pm$ 7.92 
 & 93.80 $\pm$ 0.60 & 61.06 $\pm$ 4.29 & 73.88 $\pm$ 3.10 
\\\Xhline{3\arrayrulewidth}
{BERT + {\ModelName}} & 93.60 $\pm$ 0.79&\textbf{22.03 $\pm$ 6.10}& \textbf{35.29 $\pm$ 8.05}& 93.29 $\pm$ 0.86&\textbf{66.44 $\pm$ 3.07}& \textbf{77.56 $\pm$ 2.05} \\
\Xhline{5\arrayrulewidth}
\end{tabular}
}
%\vspace{-0.1in}
\caption{Generalized few shot intent detection with 1-shot and 5-shot settings on SNIPS-NLU and NLUED. Seen is the accuracy on the seen intents ($acc_s$), Unseen/Novel is the accuracy on the novel intents ($acc_s$), H-Mean is the harmonic mean of seen and unseen accuracies.}
\vspace{-0.1in}
\label{exp}
\end{table*}

\vspace{-0.05in}
\section{Experiments}
%\vspace{-0.05in}
To evaluate the effectiveness of the proposed approach for generating labeled examples for few-shot intents, experiments are conducted for the GFSID task on two real-world datasets. The few-shot intents are augmented with utterances generated from {\ModelName}. %Both 1-shot and 5-shot settings are performed for these two datasets.

\vspace{-0.05in}
\subsection{Datasets}
%\vspace{-0.05in}
Following \cite{xia2020cg}, two public intent detection datasets are used in the experiments: SNIPS-NLU \citep{coucke2018snips} and {NLUED} \citep{XLiu.etal:IWSDS2019}. These two datasets contain utterances from users when interacting with intelligent assistants and are annotated with pre-defined intents. Dataset details are illustrated in Table \ref{dataset}.

\noindent\textbf{SNIPS-NLU\footnote{https://github.com/snipsco/nlu-benchmark/}} contains seven intents in total. Two of them (RateBook and AddToPlaylist) as regraded as few-shot intents. The others are used as existing intents with sufficient annotation. We randomly choose 80\% of the whole data as the training data and 20\% as the test data. 

\noindent\textbf{NLUED}\footnote{https://github.com/xliuhw/NLU-Evaluation-Data} is a natural language understanding dataset with 64 intents for human-robot interaction in home domain, in which 16 intents as randomly selected as the few-shot ones. A sub-corpus of 11, 036 utterances with 10-folds cross-validation splits is utilized.

%\vspace{-0.1in}
\subsection{Baselines}
%\vspace{-0.05in}
We compare the proposed model with a few-shot learning model and several data augmentation methods. 1) Prototypical Network \cite{snell2017prototypical} (PN) is a distance-based few-shot learning model. It can be extended to the GFSID task naturally by providing the prototypes for all the intents. BERT is used as the encoder for PN to provide a fair comparison. We finetune BERT together with the PN model. This variation referred to as BERT-PN+.
2) BERT. For this baseline, we over-sampled the few-shot intents by duplicating the few-shots to the maximum training examples for one class.
3) SVAE \cite{bowman2015generating} is a variational autoencoder built with LSTMs.
4) CGT \cite{hu2017toward} adds a discriminator based on SVAE to classify the sentence attributes.
5) EDA \cite{wei2019eda} uses simple data augmentations rules for language transformation. We apply three rules in the experiment, including insert, delete and swap.
6) CG-BERT \cite{xia2020cg} is the first work that combines CVAE with BERT to do few-shot text generation. 
BERT is fine-tuned with the augmented training data for these generation baselines. The whole pipelines are referred to as BERT + SVAE, BERT + CGT, BERT + EDA and BERT + CG-BERT in Table \ref{exp}. An ablation study is also provided to understand the importance of contrastive loss by removing it from {\ModelName}.

%\vspace{-0.1in}
\subsection{Implementation Details}
%\vspace{-0.05in}
Both the encoder and the decoder use six transformer layers. Pre-trained weights from BERT-base are used to initialize the embeddings and the transformer layers. The weights from the first six layers in BERT-base are used to initialize the transformer layers in the encoder and the later six layers are used to initialize the decoder.
Adam optimizer \citep{kingma2014adam} is applied for all the experiments. The margin for the contrastive loss is $0.5$ for all the settings. All the hidden dimensions used in {\ModelName} is 768. For {\ModelName}, the learning rate is 1e-5 and the batch size is 16. Each epoch has 1000 steps. Fifty examples from the training data are sampled as the validation set. The reconstruction error on the validation set is used to search for the number of training epochs in the range of [50, 75, 100]. The reported performances of {\ModelName} and the ablation of contrastive loss are both trained with 100 epochs. 

The hyperparameters for the generation process including the top index $k$ and the sampling times $s$ are chosen by evaluating the quality of the generated utterances. The quality evaluation is described in section \ref{analysis}.
We search $s$ in the list of [10, 20], and $k$ in the list of [20, 30]. We use $k=30$ and $s=20$ for BERT + {\ModelName} in NLUED, while use $k=30$ and $s=10$ for all the other experiments.
When fine-tuning BERT for the GFSID task, we fix the hyperparameters as follows: the batch size is 32, learning rate is 2e-5 and the number of the training epochs is 3.

%\vspace{-0.05in}
\subsection{Experiment Results}
%\vspace{-0.05in}
The experiment results for the generalized few-shot intent detection task are shown in Table \ref{exp}. Performance is reported for two datasets with both 1-shot and 5-shot settings. For SNIPS-NLU, the performance is calculated with the average and the standard deviation over 5 runs. The results on NLUED are reported over 10 folds.

Three metrics are used to evaluate the model performances, including the accuracy on existing many-shot intents ($acc_{m}$), the accuracy on few-shot intents ($acc_{f}$) together with their harmonic mean ($H$). As the harmonic mean of $acc_{m}$ and $acc_{f}$, $H$ is calculated as:
%\vspace{-0.07in}
\begin{equation}
%\vspace{-0.07in}
    H = 2 \times (acc_{m} \times acc_{f})/(acc_{m} + acc_{f}).
%    \vspace{-0.07in}
\end{equation}
We choose the harmonic mean as our evaluation criteria instead of the arithmetic mean because the overall results are significantly affected by the many-shot class accuracy $acc_m$ over the few-shot classes $acc_f$ in arithmetic mean \cite{xian2017zero}. Instead, the harmonic mean is high only when the accuracies on both many-shot and few-shot intents are high. Due to this discrepancy, we evaluate the harmonic mean which takes a weighted average of the many-shot and few-shot accuracy.

As illustrated in Table \ref{exp}, the proposed pipeline BERT + {\ModelName} achieves state-of-the-art performance on the accuracy for many-shot intents, few-shot intents, and their harmonic mean for the SNIPS-NLU dataset. As for the NLUED dataset, BERT + {\ModelName} outperforms all the baselines on the accuracy for few-shot intents and the harmonic mean, while achieves comparable results on many-shot intents compared with the best baseline. Since the many-shot intents have sufficient training data, the improvement mainly comes from few-shot intents with scarce annotation. For example, the accuracy for few-shot intents on NLUED with the 5-shot setting improves 5\% from the best baseline (BERT + CG-BERT). 

Compared to the few-shot learning method, {\ModelName} achieves better performance consistently in all the settings. BERT-PN+ achieves decent performance on many-shot intents while lacks the ability to provide embeddings that can be generalized from existing intents to few-shot intents.

For data augmentation baselines, {\ModelName} obtains the best performance on few-shot intents and the harmonic mean. These results demonstrate the high quality and diversity of the utterances generated form {\ModelName}.
CGT and SVAE barely improve the performance for few-shot intents. They only work well with sufficient training data. The utterances generated by these two models are almost the same as the few-shot examples.
The performance improved by EDA is also limited since it only provides simple language transformation like insert and delete.
Compared with CG-BERT that incorporates the pre-trained language model BERT, {\ModelName} further improves the ability to generate utterances for few-shot intents with composed natural language generation.

From the ablation study illustrated in Table \ref{ablation}, removing the contrastive loss decreases the accuracy for few-shot intents and the harmonic mean. It shows that the contrastive loss regularizes the generation process and contributes to the downstream classification task.

\begin{table}[ht!]
\centering
%\vspace{-0.05in}
\resizebox{\linewidth}{!}{% <-
\begin{tabular}{l|ccc}
\Xhline{3\arrayrulewidth}
& Many-shot  & Few-shot & H-Mean \\ 
& \multicolumn{3}{c}{NLUED 1-shot} \\\hline
CLANG & 93.60 $\pm$ 0.79&22.03 $\pm$ 6.10& 35.29 $\pm$ 8.05\\
{     -$\mathcal{L}_v$} &93.88 $\pm$ 0.84&	21.76 $\pm$ 6.44&	34.92 $\pm$ 8.48\\ \hline
& \multicolumn{3}{c}{NLUED 5-shot} \\\hline
CLANG & 93.29 $\pm$ 0.86&66.44 $\pm$ 3.07& 77.56 $\pm$ 2.05 \\
{     -$\mathcal{L}_v$} &	92.94 $\pm$ 0.72&	65.26 $\pm$ 2.95 &76.64 $\pm$ 2.06 \\
\Xhline{3\arrayrulewidth}
\end{tabular}
}
\vspace{-0.1in}
\caption{Ablation study for removing the contrastive loss $\mathcal{L}_v$ from {\ModelName} on NLUED.}
\vspace{-0.1in}
\label{ablation}
\end{table}

\vspace{-0.1in}
\subsection{Result Analysis}
\label{analysis}
%\vspace{-0.05in}
To further understand the proposed model, {\ModelName}, result analysis and generation quality evaluation are provided in this section. We take the fold 7 of the NLUED dataset with the 5-shot setting as an example. It contains 16 novel intents with 5 examples per intent. 

The intent in this paper is defined as a pair of a domain and an action. 
The domain or the action might be shared among the many-shot intents and the few-shot intents. The domain/action that exists in many-shot intents is named as a seen domain/action, otherwise, it is called a novel domain/action.
To analyze how well our model performs on different few-shot intents, we split few-shot intents into four types: a novel domain with a seen action (Novel$_d$), a novel action with a seen domain (Novel$_a$), both domain and action are seen (Dual$_s$), both domain and action are novel (Dual$_u$). We compare our proposed model with CG-BERT on these different types. As illustrated in Table \ref{decompose}, {\ModelName} consistently performs better than CG-BERT on all the types. The performance for intents with a seen action and a novel domain improves 20.90\%. This observation indicates that our model is better at generalizing seen actions into novel domains.

\begin{table}[ht!]
\centering
\resizebox{\linewidth}{!}{% <-
\begin{tabular}{l|ccccc}
\Xhline{3\arrayrulewidth}
 &    Total & Novel$_d$ & Novel$_a$ & Dual$_s$ &  Dual$_u$\\ \hline
 Number & 16 & 4 & 8 & 3 & 1\\
CG-BERT &   58.76\% & 47.76\% &60.43\%&  67.34\% & 63.16\%\\
 {\ModelName} &  67.88\% &68.66\% &62.58\% &75.51\%& 84.21\% \\
+Improve &  9.12\% &20.90\% &2.15\%& 8.17\% & 21.05\%\\ 
\Xhline{3\arrayrulewidth}
\end{tabular}
}
\vspace{-0.1in}
\caption{Accuracies on different types of few-shot intents.}
\vspace{-0.1in}
\label{decompose}
\end{table}

As a few-shot natural language generation model, diversity is a very important indicator for quality evaluation.
We compare the percentage of unique utterances generated by {\ModelName} with CG-BERT. In CG-BERT, the top 20 results are generated for each intent by sampling the hidden variable for once. There are 257 unique sentences out of 320 utterances (80.3\%). In {\ModelName}, the top 30 results for each intent are generated by sampling the latent variables for once. We got 479 unique sentences out of 480 utterances (99.8\%), which is much higher than CG-BERT. 

Several generation examples are shown in Table \ref{case}. {\ModelName} can generate good examples (indicated by G) that have new slots values (like time, place, or action) not existing in the few-shot examples (indicated by R). For example, G1 has a new time slot and G5 has a new action. Bad cases (indicated by B) like B1 and B5 fill in the sentence with improper slot values. {\ModelName} can also learn sentences from other intents. For instance, G3 transfer the expression in R3 from ``Recommendation Events'' to ``recommendation movies''. However, B4 fails to transfer R4 into the movie domain.

\begin{table}[hbt!]
\centering
\resizebox{\linewidth}{!}{% <-
\begin{tabular}{l}
\Xhline{3\arrayrulewidth}
Intent: \textbf{Alarm Query} \\ \hline
R1: what time is my alarm set for tomorrow morning \\ 
%R2: what am i up to this weekend \\
G1: what time is my alarm set for \textcolor{red}{this weekend}\\
B1: \textcolor{blue}{how much} my alarm set for tomorrow morning \\ \hline
R2: i need to set an alarm how many do i have set \\ 
G2: do i have \textcolor{red}{an alarm set for tomorrow morning} \\
B2: how many \textcolor{blue}{emails} i have set\\ \Xhline{3\arrayrulewidth}
Intent: \textbf{Recommendation Movies}\\ \hline
R3 (events): is there anything to do tonight \\ 
G3 (movies):\textcolor{red}{are there anything movie tonight}\\
\hline
R4 (events): what bands are playing in town this weekend \\
B4 (movies): \textcolor{blue}{what bands are playing in town this weekend} \\
\Xhline{3\arrayrulewidth}
Intent: \textbf{Takeaway Order}\\ \hline
R5: places with pizza delivery near me\\
G5: \textcolor{red}{search for the} delivery near me \\
B5: \textcolor{blue}{compose} a delivery near me\\
G6: places with pizza delivery near \textcolor{red}{my location} \\
B6: places with pizza delivery near \textcolor{blue}{my pizza} \\
 \Xhline{3\arrayrulewidth}
\end{tabular}
}
\vspace{-0.1in}
\caption{Generation examples from {\ModelName}. R are real examples in the few-shots, G are good generation examples and B are bad cases.}
\vspace{-0.1in}
\label{case}
\end{table}

Case study is further provided for the Alarm Query intent with human evaluation. There are 121 unique utterances generated in total. As shown in Table \ref{human}, 80.99\% are good examples and 19.01\% are bad cases.
Good cases mainly come from four types: Add/Delete/Replacement which provides simple data augmentation; New Time slot that has a new time slot value; New Question that queries alarm in new question words; Combination that combines two utterances together. Bad cases either come from a wrong intent (intents related to Query or Alarm) or use a wrong question word.
\begin{table}[ht!]
\centering
\resizebox{\linewidth}{!}{% <-
\begin{tabular}{l|c|c}
\Xhline{3\arrayrulewidth}
Type & Count & Percent \\ \hline
Add/Delete/Replacement  & 33 & 27.27\%\\
New Time slot &  30  & 24.79\% \\
New Question  & 28 & 23.14\%\\
Combination &  7 & 5.79\% \\\hline
Total Good Cases &98 & 80.99\%\\ \Xhline{3\arrayrulewidth}
Wrong Intent (Query)  & 10 & 8.26\% \\
Wrong Intent (Alarm) & 7 & 5.79\%\\ 
Wrong Question & 6 & 4.96\%\\ \hline
Total Bad Cases &23 & 19.01\% \\\hline
\Xhline{3\arrayrulewidth}
\end{tabular}
}
\vspace{-0.1in}
\caption{Generation case study for the Alarm Query intent.}
\vspace{-0.1in}
\label{human}
\end{table}

\vspace{-0.07in}
\section{Related Work}
\vspace{-0.03in}
\noindent\textbf{Generative Data Augmentation for SLU}
Generative data augmentation methods alleviates the problem of lacking data by creating artificial training data with generation models. Recent works \cite{wei2019eda, malandrakis2019controlled, yoo2019data} have explored this idea for SLU tasks like intent detection. 
\citet{wei2019eda} provide data augmentation ability for natural language with simple language transformation rules like insert, delete and swap. 
\citet{malandrakis2019controlled} and \citet{yoo2019data} utilize variational autoencoders \cite{kingma2013auto} to generate training data for SLU tasks. \citet{malandrakis2019controlled} investigates templated-based text generation model to augment the training data for intelligent artificial agents. \citet{yoo2019data} generate fully annotated utterances to alleviate the data scarcity issue in spoken language understanding tasks. These models utilize LSTM as encoders \cite{hochreiter1997long} with limited model capacity. \citet{xia2020cg} provide the first work that combines CVAE with BERT to generate utterances for generalized few-shot intent detection.

Recently, large-scale pre-trained language models are proposed for conditiaonal text generation tasks \cite{dathathri2019plug, keskar2019ctrl}, but they are only evaluated by human examination. They are not aiming at improving downstream classification tasks in low-resource conditions.

\noindent\textbf{Contrastive Learning in NLP} 
Contrastive learning that learns the differences between the positive data from the negative examples has been widely used in NLP \citep{gutmann2010noise, mikolov2013distributed, 2019arXiv191103047C}. \citet{gutmann2010noise} leverage the Noise Contrastive Estimation (NCE) metric to discriminate the observed data from artificially generated noise samples.
%\citet{mikolov2013distributed} proposes a negative sampling method for word representation learning.
\citet{2019arXiv191103047C} introduce contrastive learning for multi-document question generation by generating questions closely related to the positive set but far away from the negative set. Different from previous works, our contrastive loss learn a positive example against a negative example together with label information.
\vspace{-0.05in}
\section{Conclusion}
\vspace{-0.05in}
In this paper, we propose a novel model, Composed Variational Natural Language Generator ({\ModelName}) for few-shot intents. An intent is defined as a combination of a domain and an action to build connections between existing intents and few-shot intents. {\ModelName} has a bi-latent variational encoder that uses two latent variables to learn disentangled semantic features corresponding to different parts in the intent. These disentangled features are composed together to generate training examples for few-shot intents. Additionally, a contrastive loss is adopted to regularize the generation process. Experimental results on two real-world intent detection datasets show that our proposed method achieves state-of-the-art performance for GFSID.

\section*{Acknowledgments}
We thank the reviewers for their valuable comments.
This work is supported in part by NSF under grants III-1763325, III-1909323, and SaTC-1930941.

\bibliography{emnlp2020}
\bibliographystyle{acl_natbib}

\end{document}